\documentclass[11pt]{article}
\usepackage[dvipsnames]{xcolor}
\usepackage{rldmsubmit}
\usepackage{palatino}
\usepackage{graphicx}
\usepackage{microtype}
\usepackage{booktabs}
\usepackage{soul}
\usepackage{enumitem}
\usepackage[capitalise,noabbrev]{cleveref}

\title{Objective Metrics for Human-Subjects Evaluation in\\Explainable Reinforcement Learning}

\author{
Balint Gyevnar\thanks{Equal contribution} \\
School of Informatics \\
University of Edinburgh \\
\texttt{balint.gyevnar@ed.ac.uk} \\
\And
Mark Towers$^*$ \\
School of Electronics and Computer Science \\
University of Southampton \\
\texttt{mt5g17@soton.ac.uk}
}

\newcommand*\inlinegraphics[1]{\raisebox{-1.5pt}{\includegraphics[height=1em]{#1}}}

\begin{document}

\maketitle

\begin{abstract}
Explanation is a fundamentally human process. Understanding the goal and audience of the explanation is vital, yet existing work on explainable reinforcement learning (XRL) routinely does not consult humans in their evaluations. Even when they do, they routinely resort to subjective metrics, such as confidence or understanding, that can only inform researchers of users' opinions, not their practical effectiveness for a given problem. This paper calls on researchers to use objective human metrics for explanation evaluations based on observable and actionable behaviour to build more reproducible, comparable, and epistemically grounded research. To this end, we curate, describe, and compare several objective evaluation methodologies for applying explanations to debugging agent behaviour and supporting human-agent teaming, illustrating our proposed methods using a novel grid-based environment. We discuss how subjective and objective metrics complement each other to provide holistic validation and how future work needs to utilise standardised benchmarks for testing to enable greater comparisons between research.
\end{abstract}


\acknowledgements{We thank Tobias Huber and Hendrik Baier for the thought-provoking discussions. We also thank David Abel for his constructive comments.}

\startmain 

\section{Explainable Agents}
\label{sec:intro}
Most decisions are based on a wide array of causes.
The capacity to select among these causes and formulate an explanation allows humans to learn from past actions, communicate justifications, and act safely with accountability~\cite{hitchcockPortableCausalDependence2012,kirfelInferenceExplanation2022,alickeCausationNormViolation2011,lombrozoCausalExplanatoryPluralism2010}.
When our actions, partly or wholly, are ceded to autonomous agents, we tend to ascribe intentions to them as we do to humans~\cite{gyevnar2024attribute,clarkSocialRobotsDepictions2023,perez-osorioAdoptingIntentionalStance2020} and look for similar explanatory capabilities~\cite{karimiSurveyAlgorithmicRecourse2022,abbassSocialIntegrationArtificial2019}.
With the rise of `agentic systems' relying on ever-larger neural networks fine-tuned or built with RL, explainability---defined here as an agent's capability to formulate a human-intelligible explanation of its actions---is essential.

Explainability is fundamentally \emph{for} humans; thus, we should interpret and evaluate them in contexts with humans~\cite{millerExplanationArtificialIntelligence2019,ehsanChartingSociotechnicalGap2023,byrneGoodExplanationsExplainable2023,gyevnar2023transparencyGap}.
A good explanation can reveal the purpose behind actions, which, in turn, teaches people when to trust autonomous agents and when to contest their decisions.
Explanations must be clear and intelligible, as humans perceive and interpret them with limited cognitive bandwidth.
Lastly, an explanation can reveal, to the extent necessary, information to account for mistakes and trace errors.
Of course, trying to satisfy these criteria simultaneously is counterproductive because different explanations are helpful for different audiences.
For example, an explanation designed to help debug faulty behaviour will be helpful for a researcher but not for calibrating trust with end users. 
Therefore, understanding the goal and audience of an explanation is vital. 

An aspect of ensuring that explanations achieve their envisioned goal for their target audience is using the appropriate evaluation methodologies.
Using the terminology of Milani \textit{et al.}~\cite[Table 1.]{milaniExplainableReinforcementLearning2024}, there is an overwhelming preference in XRL for measuring fidelity and performance as go-to methodologies. At the same time, a persistent issue is the neglect of human-centred evaluation.
Current work that does perform human participants testing commonly favours self-reported subjective metrics of explanation quality, such as comprehensibility and preferability (see, e.g.,~\cite{hoffmanMetricsExplainableAI2019,mohseniMultidisciplinarySurveyFramework2021}).
When used appropriately, these subjective metrics are undoubtedly helpful in mapping human preferences, but researchers often apply the recommendations of existing work, particularly Hoffman \textit{et al.}'s~\cite{hoffmanMetricsExplainableAI2019}, without foundations in experimental survey design, which can inadvertently introduce modelling errors and statistical bias.
Subjective metrics may also be more sensitive to the demographics of the evaluators and the phrasing of individual questions, making for brittle experimental designs with questionable reproducibility.
Nevertheless, researchers often claim that their explanation is successful because it has performed well on these subjective metrics despite not testing on its goal and target audience.
As such, subjective metrics of explanation quality alone are insufficient for evaluating XRL.
We argue instead that understanding and measuring an explanation's \emph{actionability}---which we define as the ability of explanations to affect or change human behaviour---is just as crucial. 

Assessing actionability should be based on objective measurements of human behaviour, which, in turn, should be determined by a clearly defined set of goals for the system, its audience, and its explanations.
Such objective human metrics enable building practically applicable, comparable, and epistemically grounded research in XRL, fostering better reproducibility, incremental work, and more rigorous practices.
To these ends, the rest of this paper presents the first curation and comparison of objective metrics for human-subjects evaluation in XRL, focusing on debugging and human-agent teaming for sequential decision-making as the primary two goals of explanation.
We analyse the ramifications of using different metrics on how and what kind of explanations may be generated about autonomous agents, summarising the metrics in~\cref{tab:debugging-metrics-summary,tab:teaming-metrics-summary}.

\section{Objective Measures for XRL}
\label{sec:objective-measures}
To ground and measure an explanation's actionable effectiveness, we outline several metrics, all inspired by practical applications of XRL. We identify debugging (Section \ref{subsec:debugging}) and human-agent teaming (Section \ref{subsec:teaming}) as two primary situations in which explanations can be applied for different reasons. Before deployment of agents, developers and researchers are interested in confirming expected agent behaviour and post-deployment for fault investigations to more deeply understand why decisions were taken that went wrong. We refer to these types of applications as ``debugging''. While during deployment, collaborators (human or `agentic') use explanations for coordination, trust, or accountability, we refer to the use of explanations in this setting as ``teaming''. For both of these general applications, we suggest several measures providing definitions, measurement methodologies, and the ramifications on explanations (which explanations should be (un)acceptable, (in)efficient, or (un)trustworthy).  

\subsection{Illustrative Example}
\label{ssec:example}

\begin{figure}
    \centering
    \includegraphics[width=0.9\linewidth]{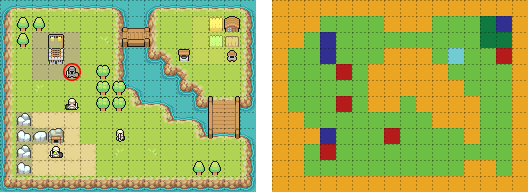}
    \caption{The mini-world environment and its semantic map. The environment has passable \textcolor{YellowGreen}{\textbf{tiles}} and impassable \textcolor{orange}{\textbf{obstacles}} such as trees, rocks, and water. There are three types of \textcolor{BrickRed}{\textbf{agents}}: \inlinegraphics{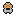} farmers, \inlinegraphics{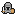} soldiers, and \inlinegraphics{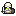} skeletons. Agents spawn at differing but fixed rates from their corresponding \textcolor{blue}{\textbf{buildings}} and can move in the four cardinal directions. The task of a farmer is to till the \textcolor{ForestGreen}{\textbf{soil}}, collect water from the \textcolor{ProcessBlue}{\textbf{well}}, water the soil, plant seeds, and then harvest the crop. The task of the soldiers is primarily to protect the farmers and kill skeletons. The skeletons' goal is to kill everyone else. In our examples, we assume that an XRL agent was trained to control the soldier highlighted with a red circle.}
    \label{fig:gridworld}
\end{figure}

We provide an easy-to-access illustration of the application of each metric using a novel grid-based environment described in detail in~\cref{fig:gridworld}, which we call \emph{mini-world}.
Each of the proposed objective metrics is then illustrated using a variation of this scenario.
We always assume that the system is controlling the same agent (i.e., the soldier in the red circle)., but for~\cref{subsec:teaming}, where we discuss human-agent teaming, we additionally assume that a human can control the sole villager in the scenario (located in the north-east of the map).

\subsection{Assessing Explanations for Debugging}
\label{subsec:debugging}
Understanding why an agent takes an action is critical both before and after deployment. Prior to releasing an autonomous agent to the real world, it is imperative to verify their decision-making, both that the system matches human expectations and acts according to laws and regulations. Additionally, fault investigations (post-deployment) are a standard procedure in engineering for understanding how and why a mistake was made. In both cases, explanations are helpful tools to achieve these purposes. We devise various evaluation methodologies from these applications summarised in Table \ref{tab:debugging-metrics-summary} and outlined below.

\begin{table}[t]
    \centering
    \caption{Summary of objective evaluation metrics for evaluating explanation debugging effectiveness.}
    \label{tab:debugging-metrics-summary}
    \begin{tabular}{@{}lp{5.25cm}p{8cm}@{}}
    \toprule 
        \textbf{Task} & \textbf{Metric} & \textbf{Pros \& Cons}  \\ 
    \midrule
        Next Action Prediction & Accuracy of predicting an agent's next $N \geq 1$ actions. &
            \textcolor{ForestGreen}{(+) Easy to verify \& versatile for all environments.} \newline 
            \textcolor{BrickRed}{(--) Difficult to implement for environments with a large number of discrete or continuous actions.} \newline
            \textcolor{BrickRed}{(--) Could be easy to guess.} \newline
            \textcolor{BrickRed}{(--) Small explanatory value with long horizons.} \vspace{1em} \\
        Goal prediction & Accuracy of selecting the correct overall agent goal from a list of candidate goals. & 
            \textcolor{ForestGreen}{(+) Robust to variations in the environment.} \newline 
            \textcolor{ForestGreen}{(+) Easy to measure.} \newline 
            \textcolor{BrickRed}{(--) Requires several reward functions with descriptions and trained agents.} \vspace{1em} \\
        Sub-Goal Prediction & Accuracy of predicting the next sub-goal of an agent. & 
            \textcolor{ForestGreen}{(+) Large explanatory value with long horizons.} \newline 
            \textcolor{BrickRed}{(--) No universally agreed definition of a sub-goal.} \vspace{1em} \\
        Counterfactual Policy & Identifying the state changes for an agent to choose any other or a particular action. & 
            \textcolor{ForestGreen}{(+) Easy to verify \& versatile for all environments.} \newline 
            \textcolor{BrickRed}{(--) The minimum state changes is computational expensive to verify.} \vspace{1em} \\
        Time Taken & How long survey participants take to answer a question & 
            \textcolor{ForestGreen}{(+) Easy to measure.} \newline 
            \textcolor{BrickRed}{(--) Very noisy.} \\
    \bottomrule
    \end{tabular}
\end{table}

\paragraph{Next Action Prediction} One immediate reason users might ask about an agent's decision-making is ``Why did you take action A?'' This question can be interpreted in various forms, including the combination of features that resulted in the action, the counterfactual rationale for why other feasible actions were not taken, etc. Researchers can test if users can follow an explanation for a given state to reach the same action as an agent. More formally, with a policy, $\pi$, a state $s$, and an explanation $E$ of the policy for the state, can users predict an agent's action, $\pi(s)$ from a set of possible actions?~\footnote{For environments with continuous or very large discrete action spaces, it is infeasible to present all actions; instead, a subset of actions could be presented.} This measures user accuracy in interpreting a policy's decision-making for state $s$, for example, in Madumal \textit{et al.}~\cite{madumal2019explainablereinforcementlearningcausal}. This can be extended to assess a user's ability to predict the subsequent $N$ actions or predict what an agent believes is the worst action to take in a situation, providing a more holistic evaluation. Despite its ease of implementation, just needing to sample an agent's policy for a suite of states, its real-world applicability might be minimal. Agents commonly take hundreds or thousands of actions to achieve a goal, meaning that researchers would need to review hundreds of explanations and might struggle to interpret an agent's larger decision-making rationale from the explanation. Further, such evaluations prevent explanations from including the agent's next action, a beneficial or necessary feature for some. 

\begin{description}[font=\normalfont,leftmargin=1em]
    \item[\textit{Example.}] Some explanation mechanism teaches the participant to select the correct action of the soldier by showing visualisations of the Q-values of its actions in various scenarios. During evaluation in the example scenario, the participant needs to pick what action the soldier will take next by selecting one from a list of available actions (e.g., north, south, east, west, attack) or possibly by performing the next action in an interactive version of the environment. In this example, the soldier moves north to protect the villager from the eastmost skeleton threatening it. 
\end{description}

\paragraph{Goal Prediction} At the opposite explanatory temporal scale to Next Action Prediction, users can ask, ``What is the agent working to achieve?''. We interpret this as referring to an agent's top-level goal that can generally be summarised in natural language. For agents, they optimised an environment's reward function, $R$, to maximise their cumulative reward, i.e., their goal. Importantly, researchers can modify the reward function for most environments so that an agent achieves alternative goals. This can include irrational goals (to humans), non-standard goals, or the individual sub-goals that compose an agent's global goal. This can be developed into an evaluation methodology through implementing $N$ separate, non-overlapping reward functions, $R_1, R_2, \dots, R_N$, with a unique policy learnt for each, $\pi_{R_1}, \pi_{R_2}, \dots, \pi_{R_N}$. Like the Next Action Prediction, users are tasked with determining an agent's particular goal, $\pi_{R_i}$, trained to optimise $R_i$, for a given environment state, $s$ and explanation, $E$, from a list of possible goals described in natural language. For example, Huber \textit{et al.}~\cite{huber2023ganterfactualrlunderstandingreinforcementlearning} asks users to suggest an agent's strategy. This methodology encourages explanations highlighting an agent's global or higher-level thinking beyond state $s$ and can be tested in parallel with Next Action Prediction. However, like Next Action Prediction, explanations that are evaluated this way are partially limited from including prior knowledge of the complete goal description, as that would defeat the purpose of the metric.

\begin{description}[font=\normalfont,leftmargin=1em]
    \item[\textit{Example.}] Some explanation mechanism teaches the participant to infer the agents' goal by describing the expected rewards for various rolled out state sequences that end in a terminal state. During evaluation, the participant needs to pick from high-level descriptions of different goals for the soldier, some of which may seem irrational to the participant (e.g., protect the villager, only kill skeletons, die as soon as possible). In this example, the soldier behaves as expected and prioritises protecting the villager over killing skeletons.
\end{description}

\paragraph{Sub-Goal Prediction} For episodes that are hundreds or thousands of actions long, the next action and final goals are two temporal extremes to explain an agent's decision-making. An intermediate question of an agent's decision-making might be, ``What is the agent's goal for the next handful of actions?''. These intermediate or sub-goals are iteratively completed to achieve an agent's overall goal, making them explanatory and evaluatory targets. However, objective evaluation with sub-goals is less straightforward than the Next Action or Goal prediction, as we are unaware of an objective method for determining an agent's ``true'' sub-goal. Instead, researchers might need to handpick and annotate states with a reasonable and consistent sub-goal following agent behaviour.~\footnote{There are several approaches for this problem; however, their effectiveness for this application is unknown.} The advantage of Sub-Goal Prediction is its applicability to debugging real-world sequential decision-making problems where developers wish to understand an agent's short-term beliefs or decision-making. Further, sub-goals are not biased by an explanation's knowledge of an agent's next action or final goal, expanding the space of possible evaluable explanations. 

\begin{description}[font=\normalfont,leftmargin=1em]
    \item[\textit{Example.}] Some explanation mechanism teaches the participant to infer the agents' sub-goals by describing the expected rewards for various rolled out partial state sequences. During evaluation, the participant needs to pick from high-level descriptions of different sub-goals for the soldier, some of which may seem irrational to the participant (e.g., go over the nearest bridge, kill the skeleton in front of you, go to the skeletons' cave). In this example, the soldier behaves as expected and prioritises protecting the villager over killing skeletons so it moves over the nearest bridge.
\end{description}

\paragraph{Counterfactual Policy} It can often not be enough to comprehend an agent's decision-making for action but understand what must change in the state to influence the optimal action. We propose two evaluation variations to predict the state changes, $s^\prime$, that lead the policy to a different action from the original state, $\pi(s) \neq \pi(s^\prime)$ or a particular action $A$, $\pi(s^\prime) = A$. These evaluation methodologies can be implemented, for example, in Gyevnar \textit{et al.}~\cite{gyevnar2024causal}, by either showing users a set of states and selecting the one that minimises the state differences or using interactive software where users modify the state to fulfil the criteria. Like Next Action Prediction, interpretable policies such as decision trees are advantageous as users can mentally simulate an state to check if the policy action changes compared to neural networks-based policies. However, for counterfactual explanation mechanisms, it is not easy to assess them with this evaluation as it would be common to present the answer as part of the explanations, eliminating any cognitive requirements for users.   

\begin{description}[font=\normalfont,leftmargin=1em]
    \item[\textit{Example.}] Some explanation mechanism teaches the participant about the most important environmental elements that influenced the decisions of the soldier. During evaluation, the participant needs to pick from a list of pre-determined changes to the environment (e.g., remove nearest skeleton, remove farther bridge, change position of villager, etc.) or possibly by performing a change themselves in an interactive version of the environment. In this example, moving the villager three tiles nearer to the soldier makes it easier for the soldier to protect it and it alters the eastmost skeletons behaviour to move north-west so that the soldier now has time to attack the nearest skeleton.
\end{description}

\paragraph{Time Taken} The cognitive complexity and load on users for an explanation mechanism is important to measure. Subjective questions can partially ascertain this, but users' time to answer a question provides a proxy metric. Importantly, this can be measured in addition to the primary evaluation methodology, e.g., Next Action, Goal, and/or Sub-Goal Prediction. Measuring the time taken for different explanation mechanisms should encourage researchers to develop explanations that users can easily interpret and comprehend the answer. Further, for the average and individual question time taken, researchers can investigate if the accuracy of different mechanisms is correlated with the average time taken or if correct and incorrect answers correlate with the time taken.

\subsection{Assessing Explanations in Human-Agent Teaming}
\label{subsec:teaming}
More underdeveloped and currently less appreciated are the applications of explanations in agentic systems where human-agent or agent-agent teaming is important. To support these systems, explanations are anticipated to justify the decisions of both agent and human and coordinate information between users. For agent-agent teaming, it is feasible that these explanations are not humanly interpretable. Still, we believe that making such communication understandable by humans could allow greater oversight and the ad-hoc incorporation of humans into such systems. These teaming situations can exist in several formulations: \emph{oversight}, where humans observe or instruct agent(s) to complete a task requiring bi-directional communication; \emph{support}, where human(s) complete a task and an agent(s) assists; and \emph{cooperation}, where both agent(s) and human(s) operate together within an environment requiring collaboration. Like debugging in the previous section, assessing the effectiveness of explanations in teaming settings is critical to understanding agents' strengths and weaknesses.  

\begin{table}[t]
    \centering
    \caption{Summary of objective evaluation metrics for evaluating explanation effectiveness in teaming situations.}
    \label{tab:teaming-metrics-summary}
    \begin{tabular}{@{}lp{6cm}p{8cm}@{}}
    \toprule \textbf{Task} & \textbf{Metric} & \textbf{Pros \& Cons}  \\ \midrule
        Task Competition & The score or reward achieved by agents or humans completing a task. & 
            \textcolor{ForestGreen}{(+) Generally easy to measure.} \newline 
            \textcolor{BrickRed}{(--) Can be achieved without an explanation.} \vspace{1em} \\
        Inter-Agent Conflict & The amount of disagreement between agents. & 
            \textcolor{ForestGreen}{(+) A practical real-world problem.} \newline 
            \textcolor{BrickRed}{(--) Difficult to measure.} \vspace{1em} \\
        Time Taken & How long survey participants take to answer a question. & 
            \textcolor{ForestGreen}{(+) Easy to measure.} \newline 
            \textcolor{BrickRed}{(--) Very noisy.} \newline
            \textcolor{BrickRed}{(--) Isn't necessarily correlated with effectiveness.} \\
    \bottomrule
    \end{tabular}
\end{table}

\paragraph{Task Completion} In teaming scenarios, agents work together to complete a task, communicating and coordinating. While not necessary for all tasks, explanations can be a core component of teamwork and capability for effectively completing novel or complex tasks. To implement this, agents can be scored on their cumulative quantity or quality of work achieved across various teaming situations. An example is Puri \textit{et al.}~\cite{puri2020explainmoveunderstandingagent}, who measured human performance when solving chess puzzles with a support-agent providing explanatory help. This enables the comparison of human-agent capabilities using explanation as a communication channel. 

\begin{description}[font=\normalfont,leftmargin=1em]
    \item[\textit{Example.}] An explanation mechanism notifies the human (the villager) of the agent's intention to cross the bridge, because there is a skeleton approaching from the south. In reaction, the human can move closer to the soldier, explaining to the agent that this enables it to attack the nearest skeletons while still allowing the human to perform their farming task. The repeated use of these explanations should then lead to a higher final score at the end of the episode.
\end{description}

\paragraph{Inter-Agent Conflict} When collaborating, agents can disrupt each other while attempting to solve the task. Therefore, coordination is a feasible application for explanations to manage agent priorities and promote problem-solving. We propose measuring the number of situations where agents fail to communicate accurately, attempt to complete the same sub-task or problem, or block/slow down each other's movements. We refer to these situations as Inter-Agent Conflicts, a crucial feature that may reduce the deployment effectiveness of humans and agents solving problems together. Implementing such metrics will depend on the environment and the mode of deployment of agents, but measuring this metric can inform the design of explanations and agents that address the practical complexities of teamwork in the real-world. 

\begin{description}[font=\normalfont,leftmargin=1em]
    \item[\textit{Example.}] In the example scenario, the villager may be controlled by another agent. In this collaborative task, without an explanation, the agents might end of staying next to each other, so that the soldier can always defend the villager. However, this is greatly suboptimal, as the villager may not be able to farm. Instead, similarly to the previous example, explanations could be used to coordinate information between the two agents to avoid conflict.
\end{description}

\paragraph{Time Taken} For almost all teaming situations, assuming the same performance, the faster a task is completed, the better. Like debugging, measuring the time teams take to complete a task or problem can provide insight into the cognitive complexity of explanations and how users interact with them. 

\section{Discussion}
\label{sec:discussion}
This paper presents a first step towards the objective human testing of XRL algorithms which, in turn, aims to foster the creation of more practical, comparable, and epistemically grounded research with better reproducibility and an opportunity to build on and reproduce other researchers' work.
Remembering that understanding the goal and audience of an explanation is vital, we curate and compare a range of objective metrics for XRL within two goals: debugging and teaming (\cref{subsec:debugging,subsec:teaming}, respectively). 
For both, we outline collection methodologies, the ramifications on explanation mechanisms, and how they limit the space of testable explanation mechanisms.

While focusing on objective metrics, we recognise the value and importance of including self-reported subjective questions with human participant studies. Although they may only qualify a user's preference for an explanation mechanism, not its actionability, understanding how users subjectively view an explanation is essential. With users' reported confidence, understanding, and similar questions, researchers can investigate whether there are correlations between objective and subjective questions, and the tendency for adoption. For example, ``Is the user's debugging accuracy correlated with their confidence?'' or ``Does an explanation helpfulness correlate with a human-agent team's speed or accuracy in solving a problem?'' Importantly, researchers should desire user beliefs to match the explanation's performance. Otherwise, users might have strong incorrect beliefs about an agent or environment, which can further mislead or mistake users than those without explanations. 

We also believe that there is a pronounced need for insights from cognitive psychology, which has a long track record of analysing and characterising the human mechanisms of explanation. Importantly, in practice, our proposed metrics may be instantiated in a wide range of ways and their effectiveness in capturing explanation success will vary on the quality of the experimental setup. Cognitive science has a lot to teach XRL in regards to how one should control the experimental variables such that our metrics measure what they claim to.

Further, we recommend that researchers add a ``No Explanation'' in any study to provide a baseline to check if user performance is close to random or biased in certain cases. In some cases, this No Explanation could be a natural language description of an agent's next action to see if users can effectively use this information without using an agent's decision-making. In fact, several previous works have found that people often offer an explanation of what others might describe as a description~\cite{fuentes2024computational}, which serves as additional motivation to a No Explanation approach.

Within the context of human participants' testing of XRL, we note the importance of environment selection as the testing ground for explanations. We believe that while investigating with toy (e.g., CartPole) or gridworld environments is a necessary first step, the field should strive to understand explanations in more complex environments that are closer to their true applications or deployments. This should help ensure that the field develops algorithms that adequately scale to the expected complexity of agent behaviour and decision-making in the real-world. Setting appropriate ``grand challenges'' in the form of, for example, a competition, could serve not only to expand the scope of the field but also to increase its public-facing profile, much like how AlphaGo fuelled interest in RL and deep learning.

Finally, another limitation of XRL not discussed in this paper is the lack of standardised benchmarks, which prevent comparisons between research even if objective metrics are used. We hope that with this suite of comparable metrics, future research can utilise common evaluation and use cases to enable better comparisons and understanding of explanation mechanisms' strengths and weaknesses.

\bibliographystyle{abbrv}
\bibliography{refs}

\end{document}